\documentclass[runningheads]{llncs}
\usepackage[T1]{fontenc}
\usepackage{graphicx}
\usepackage{multicol}
\usepackage{multirow}
\usepackage{amsmath}
\usepackage{longtable}
\usepackage{array}
\usepackage{verbatim}
\begin{document}
\title{Enhancing Authorship Attribution through Embedding Fusion: A Novel Approach with Masked and Encoder-Decoder Language Models}
%
%
\author{Arjun Ramesh Kaushik \and Sunil Rufus R P\and
Nalini Ratha}
%
%
\institute{University at Buffalo, The State University of New York, USA 
\email{\{kaushik3,sunilruf,nratha\}@buffalo.edu}}
\maketitle              
\begin{abstract}
The increasing prevalence of AI-generated content alongside human-written text underscores the need for reliable discrimination methods. To address this challenge, we propose a novel framework with textual embeddings from Pre-trained Language Models (PLMs) to distinguish AI-generated and human-authored text. Our approach utilizes Embedding Fusion to integrate semantic information from multiple Language Models, harnessing their complementary strengths to enhance performance. Through extensive evaluation across publicly available diverse datasets, our proposed approach demonstrates strong performance, achieving classification accuracy greater than 96\% and a Matthews Correlation Coefficient (MCC) greater than 0.93. This evaluation is conducted on a balanced dataset of texts generated from five well-known Large Language Models (LLMs), highlighting the effectiveness and robustness of our novel methodology.
\end{abstract}
\section{Introduction}
In recent years, the landscape of natural language generation (NLG) technology has undergone remarkable advancements, revolutionizing the diversity, control, and quality of texts generated by Large Language Models (LLMs). Notably, OpenAI's ChatGPT, Google's Gemini, and Meta's Llama stand out as prime examples, showcasing exceptional performance across a myriad of tasks, including answering questions, composing emails, essays, and even code snippets. However, while these advancements herald a new era of human-like text generation at unprecedented efficiency, they also bring to the forefront pressing concerns regarding the detection and mitigation of potential misuse of LLMs. While there are many issues with LLMs in terms of their hallucinating responses and toxic language, the newfound capability of LLMs to emulate human-like text raises significant apprehensions about their potential misuse in activities in many areas such as phishing, disinformation campaigns, and academic dishonesty. Instances abound where educational institutions have resorted to banning ChatGPT due to apprehensions regarding its potential for facilitating cheating in assignments \cite{intro_gptban,intro_gptban2}, while media outlets have sounded the alarm over the proliferation of fake news generated by LLMs \cite{intro_fakenews}. Such concerns surrounding the misuse of LLMs have cast a shadow over their application in critical domains such as media and education.

Accurate detection of LLM-generated texts emerges as a pivotal requirement for realizing the full potential of NLG technology while mitigating the potentially serious consequences associated with its misuse. From the perspective of end-users, the ability to discern between human-authored and LLM-generated text holds the promise of bolstering trust in NLG systems and fostering wider adoption. For developers and researchers in the realm of Machine Learning, effective text detection mechanisms can aid in tracking generated texts and thwarting unauthorized usage.

\begin{figure*}[!ht]
    \centering
    \includegraphics[width=1\linewidth]{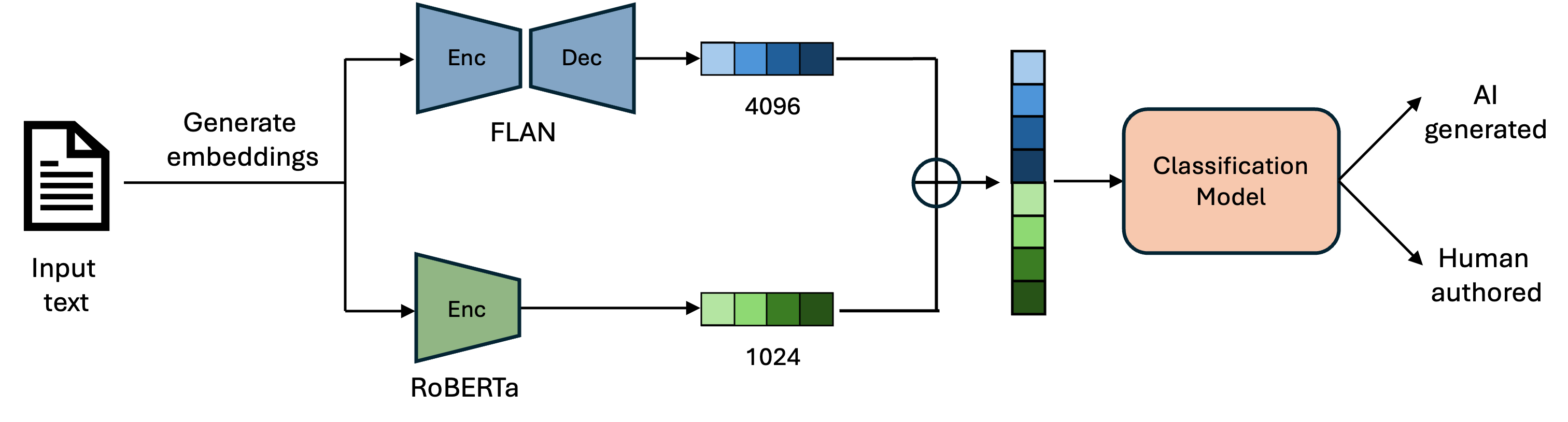}
    \caption{An overview of our framework to perform authorship attribution between machine-generated and human-authored texts.}
    \label{fig:intro_emb}
\end{figure*}

Given the critical significance of accurate LLM-generated text detection, we propose a novel framework based on representing text as images through Embedding Fusion as in Fig. \ref{fig:intro_emb}. Embeddings are low-dimensional representations of high-dimensional inputs like text, aiming to capture similarities by positioning related inputs closely in the embedding space, thereby enabling AI systems to comprehend inputs akin to human understanding. Inspired by \cite{hate_speech_emb_fusion}, we combine textual representations across Pre-trained Language Models (PLMs) and leverage their complementary strengths. Additionally, post-concatenation of the feature vectors, we reshape the fused feature vector into a 2D representation to capture the inter-embedding relationships better. Amongst the 3 types of PLMs - Autoregressive Language Models (ALMs), Masked Language Models (MLMs), and Encoder-Decoder Language Models (EDLMs) \cite{plm_survey}, our experiments indicate that the combination of the former 2 PLMs works best. We believe this is because ALMs are trained with a focus on content generation and more emphasis is placed on the final tokens. With thorough experimentation using publicly available datasets, we validate that our framework achieves better results than the state-of-the-art (SOTA) methods for the task of discerning AI-generated and human-authored text.

The rest of the paper is structured as follows. We dive into the past and recent research contributions in solving the authorship attribution task in Section 2. A detailed overview of the dataset is shown in Section 3. Sections 4 and 5 provide the methodology and results in support of our approach.

\section{Related Work}

The concept of authorship attribution using BERT embeddings has been effectively demonstrated by many researchers. PART \cite{huertastato2022part} and BertAA \cite{fabien-etal-2020-bertaa} shows how BERT embeddings can be used to grasp authors' writing styles and generate stylometric representations. \cite{saedi2021siamese} explores Siamese Networks for authorship attribution (AA), comparing their effectiveness with BERT fine-tuning. On the other hand, our research utilizes 1024-sized BERT embeddings as an image of size (32,32).

Previous works such as \cite{islam2023distinguishing} \cite{cgr_base}\cite{traditional_ml2}, use traditional Machine Learning algorithms for classification on datasets of limited scope - comparison between human-authors or consider just a single LLM. \cite{ruder2016characterlevel} discusses the application of Convolutional Neural Networks (CNNs) in character-level signal processing, serving as a motivation for our work.  

The studies \cite{ma2023ai} and \cite{tang2023science} explore the detection and regulation of AI-generated text, particularly in the context of scientific writing. In \cite{ma2023ai}, a dual approach is introduced, employing feature-based methods to categorize aspects like writing style and coherence, alongside neural network-based fine-tuning of a GPT-2 output detector model using RoBERTa, achieving a 94.6\% F1 score. Conversely, \cite{tang2023science} provides a broader overview of detection techniques, encompassing black-box methods relying on API-level access and deep learning approaches involving fine-tuning LLMs like RoBERTa. The latter study also discusses white box methods, including post-hoc rule-based and neural-based approaches, as well as inference-time watermarking techniques for modifying word selection during text generation. Together, these studies contribute to understanding and regulating AI-generated text in scientific writing, offering insights into both specific detection methodologies and broader frameworks for control and regulation.

\section{Dataset}

In this research, we consider two datasets - Human vs LLM text \cite{Kaggle_63} and Deepfake text detection \cite{deepfake}. In this paper, we are focused on solving a binary classification task, and a highly imbalanced dataset will produced biased results. \textbf{Hence, we balance the dataset by randomly sampling $n$ texts from human-generated texts, where $n$ is equal to the number of LLM-generated (GPT / Llama / FLAN / Mistral / OPT) texts in consideration.}

The Human vs LLM text dataset is a compilation of texts generated from 63 different LLMs. From the 63 LLMs, we pick texts generated from 5 different LLMs for our experiments and group variants of each LLM into a single category as in Table 1. We ensure that our dataset is always balanced.

\begin{table*}
    \centering
    \label{Kaggle_dataset}
    \caption{Distribution of Kaggle's Human vs Text corpus.}
    \begin{tabular}{|c|c|c|p{2cm}|p{2cm}|}
        \hline
        \textbf{Source} & \textbf{Variants} & \textbf{No. of Samples} & \textbf{Min. word count per sample} & \textbf{Max. word count per sample}\\ 
        \hline
         Human & - & 347,692 & \centering25 & \centering\arraybackslash 71,543\\
         \hline
         \multirow{5}{*}{FLAN} & FLAN-T5-Base & \multirow{5}{*}{45,608} &  \multirow{5}{2cm}{\hfil \hfil 25} & \multirow{5}{2cm}{\centering\arraybackslash 905}\\
         & FLAN-T5-Large & & & \\
         & FLAN-T5-Small & & & \\
         &FLAN-T5-XXL & & & \\
         &FLAN-T5-XL & & & \\
         \hline
         \multirow{4}{*}{GPT} & GPT-3.5 & \multirow{4}{*}{75,599} &  \multirow{4}{2cm}{\hfil \hfil 25} & \multirow{4}{2cm}{\centering\arraybackslash 3,565}\\
         & GPT-4 & & & \\
         & GPT-J & & & \\
         & GPT-NeoX & & & \\
         \hline
         \multirow{6}{*}{Llama} & Llama-30B & \multirow{6}{*}{42,623} & \multirow{6}{2cm}{\hfil \hfil 25} & \multirow{6}{2cm}{ \centering\arraybackslash 1,770}\\
         & Llama-65B & & & \\
         & Llama-13B & & & \\
         & Llama-7B & & & \\
         & Llama-2-70B & & & \\
         & Llama-2-7B & & & \\
         \hline
         \multirow{7}{*}{OPT} & OPT-1.3B & \multirow{7}{*}{80,151} & \multirow{7}{2cm}{\hfil \hfil  25} & \multirow{7}{2cm}{\centering\arraybackslash1,044} \\
        & OPT-30B & & & \\
        & OPT-2.7B & & & \\
        & OPT-6.7B & & & \\
        & OPT-125M & & & \\
        & OPT-350M & & & \\
        & OPT-13B & & & \\
         \hline
         \multirow{2}{*}{Mistral} & Mistral-7B & \multirow{2}{*}{10,813} & \multirow{2}{2cm}{\centering 25} & \multirow{2}{2cm}{\centering\arraybackslash21,734} \\
         & Mistral-7B-OpenOrca & & &\\ 
         \hline
    \end{tabular}
\end{table*}

The Deepfake text detection dataset was created by considering 10 datasets covering a wide range of writing tasks (e.g., story generation, news writing, and scientific writing) from diverse sources (e.g., Reddit posts and BBC news). 27 LLMs were employed for the construction of deepfake texts, resulting in a dataset of 447,674 instances in total. The dataset also comprised of domain-specific data from various domains, including opinion statements, news article writing, question answering, story generation, commonsense reasoning, knowledge illustration, and scientific writing.

For this study, we only considered knowledge illustration generated using 1000 Wikipedia paragraphs from the SQuAD context. Among the 27 LLMs, we have considered FLAN, GPT, Llama and OPT in this study. We have consolidated the text outputs from various sizes of the aforementioned LLMs, such as Llama 7B, 13B, 30B, and 65B, into a single category labeled "Llama". The SQuAD domain consisted of 20,950 human-generated texts and 26,714 machine-generated texts. We split this into a balanced set with different types of LLMs and an equal amount of human-authored texts, shown in Table \ref{deepfake table}.

       

\begin{table*}
    \centering
    \caption{Distribution of Deepfake text Detection for the SQuAD domain.}
    \begin{tabular}{|c|c|p{1cm}|p{2cm}|p{2cm}|p{2.9cm}|}
        \hline
        \textbf{Source} & \textbf{Variants} & \centering\textbf{No. of Samples} & \centering\textbf{Average Document Length} & \centering\textbf{Average Sentence Length} & \textbf{Average \# Sentences per Document}\\ 
        \hline
         Human & - & \centering6477 &  \centering232.02 & \centering\arraybackslash 18.90 & \centering\arraybackslash13.48\\
         \hline
         FLAN & FLAN-T5-Base & \multirow{5}{1cm}{\centering3885} &  \multirow{5}{2cm}{\centering 279.99} & \multirow{5}{2cm}{\centering\arraybackslash 18.80} & \multirow{5}{2.9cm}{\centering\arraybackslash 15.33}\\
         & FLAN-T5-Large & & & &\\
         & FLAN-T5-Small & & & &\\
         & FLAN-T5-XXL & & & &\\
         & FLAN-T5-XL & & & &\\
         \hline
         GPT & GPT-3.5 & \multirow{3}{1cm}{\centering1830} &  \multirow{3}{2cm}{\centering 279.99} & \multirow{3}{2cm}{\centering\arraybackslash 18.80} & \multirow{3}{2.9cm}{\centering\arraybackslash 15.33}\\
         & GPT-J & & & &\\
         & GPT-NeoX & & & &\\
         \hline
         Llama & Llama-7B & \multirow{4}{1cm}{\centering3102} & \multirow{4}{2cm}{\centering 279.99} & \multirow{4}{2cm}{\centering\arraybackslash 18.80} & \multirow{4}{2.9cm}{\centering\arraybackslash 15.33}\\
         & Llama-13B & & & &\\
         & Llama-30B & & & &\\
         & Llama-65B & & & &\\
         \hline
         OPT & OPT-125M & \multirow{9}{1cm}{\centering6477} & \multirow{9}{2cm}{\centering 279.99} & \multirow{9}{2cm}{\centering\arraybackslash 18.80} & \multirow{9}{2.9cm}{\centering\arraybackslash 15.33}\\
         & OPT-350M & & & &\\
         & OPT-1.3B & & & &\\
         & OPT-2.7B & & & &\\
         & OPT-6.7B & & & &\\
         & OPT-13B & & & &\\
         & OPT-30B & & & &\\
         & OPT-IML-1.3B & & & &\\
         & OPT-IML-30B & & & &\\
         \hline
    \end{tabular}
    \label{deepfake table}
\end{table*}

\section{Proposed Methodology}

\begin{figure*}[!ht]
    \centering
    \includegraphics[width=1\linewidth]{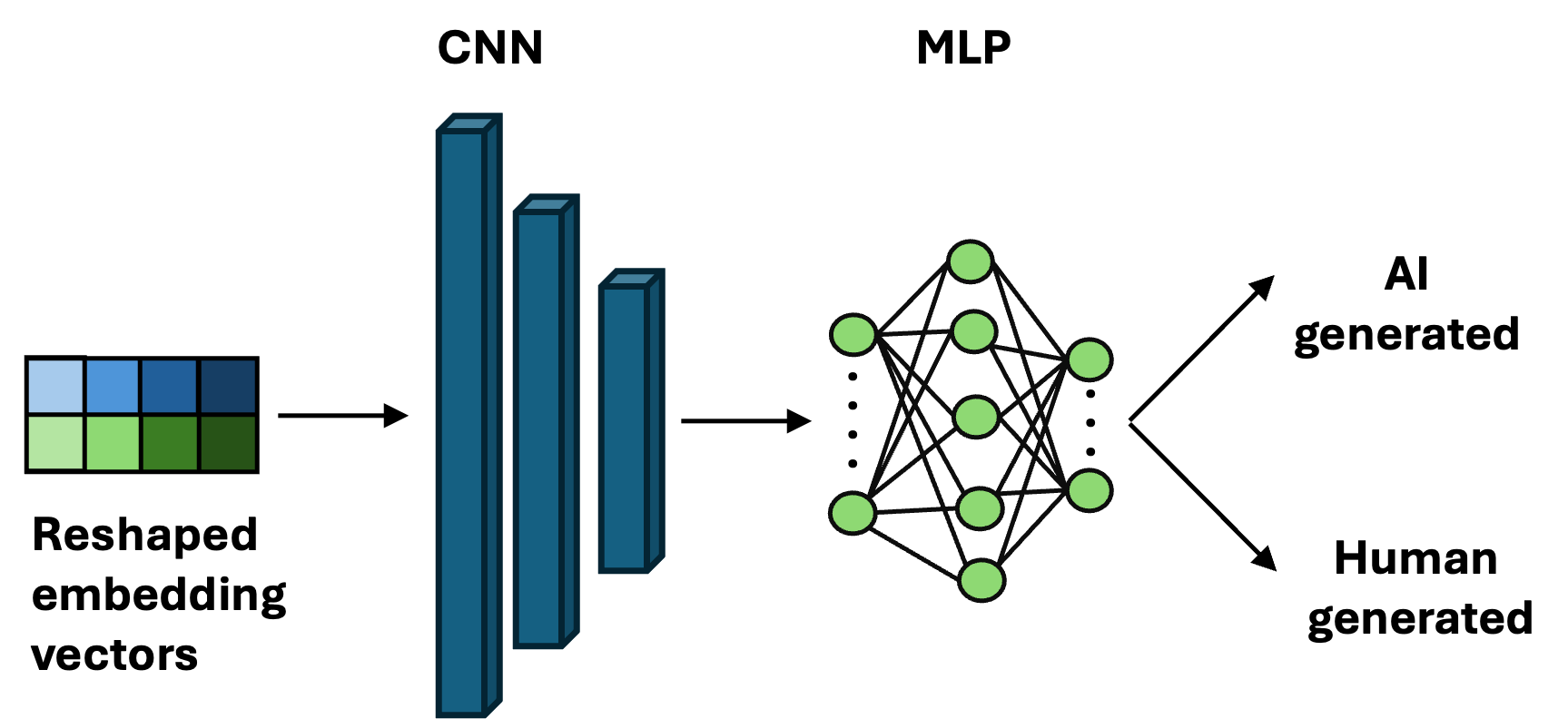}
    \caption{Architecture of the classification model used in our framework.}
    \label{fig:model}
\end{figure*}

After generating the text embeddings, we reshape the 1D embedding vectors into a 2D representation before proceeding with the classification task. This reshaping is aimed at enhancing the capture of inter-embedding relationships. The neural network architecture comprises of 3 CNN layers followed by 1 fully connected layer, as depicted in Fig.\ref{fig:model}. During training, we employ a batch size of 256, a learning rate of 0.001, and utilize the Adam optimizer.

\subsection{Autoregressive Language Models}

\begin{figure}[!ht]
    \centering
    \includegraphics[width=1\linewidth]{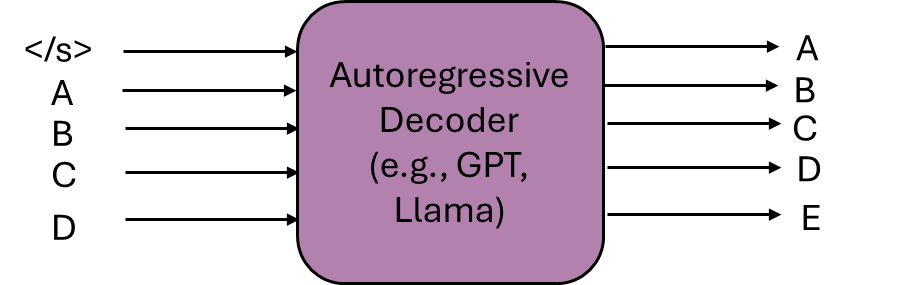}
    \caption{Autoregressive Language Models use the provided context to generate relevant responses. Here, we assume that the token 'C' captures the context of the query.}
    \label{fig:alms}
\end{figure}

An Autoregressive Language Model, as in Fig. \ref{fig:alms}, is trained to predict the succeeding word \( x_i \) based on all preceding words \( x_1, x_2, ..., x_{i-1} \). The training objective involves maximizing the log-likelihood \( \sum_i \log(P(x_i | x_1, x_2, ..., x_{i-1}; \theta_T)) \), where \( \theta_T \) represents the model parameters. In Transformer decoders, these parameters reside across multiple layers of multi-head self-attention modules. Well-known models that adhere to this architecture include GPT2 \cite{gpt2} and Llama2 \cite{llama2}. The semantic information captured by GPT2 revolves around the generation of coherent and contextually appropriate text sequences. Its training involves learning the relationships between tokens in a given context and leveraging this knowledge to generate text that follows the expected language patterns.
\subsection{Masked Language Models}
\begin{figure}[!ht]
    \centering
    \includegraphics[width=1\linewidth]{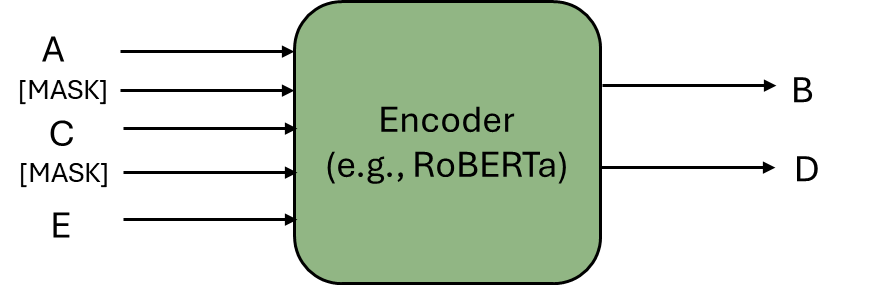}
    \caption{Masked Language Models are trained to predict masked tokens (represented as [MASK]) in a query, which helps the models better understand semantic context.}
    \label{fig:mlms}
\end{figure}

In contrast to ALMs, Masked Language Models (MLMs) as in Fig. \ref{fig:mlms} predict a "masked" word conditioned on all other words in the sequence. During MLM training, words are randomly selected to be masked, represented by a special token [MASK], or substituted with a random token. This strategy compels the model to gather bidirectional context for making predictions. The training objective is to accurately predict the original tokens at the masked positions, expressed as \( \sum_i m_i \log(P(x_i | x_1, ..., x_{i-1}, x_{i+1}, ..., x_n); \theta_T) \), where \( m_i \in \{0, 1\} \) denotes whether \( x_i \) is masked or not, and \( \theta_T \) represents the parameters in a Transformer encoder. Notably, in models like BERT, it's common practice to mask multiple words simultaneously to facilitate parallel training. Prominent examples of MLMs include BERT \cite{huertastato2022part} and RoBERTa \cite{Roberta}. RoBERTa \cite{Roberta}, an enhanced framework of BERT, excels in capturing both semantic meaning and stylometric features from textual data. These embeddings can be further utilized in various downstream tasks, such as authorship attribution, by feeding them into classification models \cite{saedi2021siamese}.

\begin{table*}
\centering
\caption{Experiments on the Human vs LLM dataset assessed the accuracies of various techniques, discerning LLM ('Source') texts solely against human-authored texts.}
\label{results_63LLMs}
\begin{tabular}{|c|p{1.8cm}|p{1.5cm}|p{1.5cm}|c|c|c|c|}
\hline
\multirow{2}{*}{\textbf{Source}} &  \multicolumn{3}{|c|}{\textbf{Embedding Model}} & \multirow{2}{*}{\textbf{Accuracy (\%)}} & \multirow{2}{*}{\textbf{TPR}} & \multirow{2}{*}{\textbf{FPR}} & \multirow{2}{*}{\textbf{MCC}}\\
\cline{2-4}
 & \centering \textbf{Encoder} & \centering \textbf{Decoder} & \textbf{Encoder-Decoder} & & & & \\
 \hline

\multirow{9}{*}{FLAN} & \centering RoBERTa & \centering - & \centering - & 94.75 & 0.923 & 0.033 & 0.891 \\
& \centering - & \centering GPT2 & \centering - & 93.06 & 0.972 & 0.111 & 0.874 \\
& \centering - & \centering Llama & \centering - & 96.10& 0.983 & 0.061 & 0.923 \\
& \centering - & \centering - & \centering FLAN & 97.02 & 0.973 & 0.032 & 0.940 \\
& \centering RoBERTa & \centering GPT2 & \centering - & 95.55 & 0.980 & 0.069 & 0.912 \\
& \centering RoBERTa & \centering Llama & \centering - & 97.09 & 0.979 & 0.038 & 0.942 \\
& \centering RoBERTa & \centering GPT2 & \centering FLAN & 97.59 & 0.986 & 0.034 & 0.952 \\
& \centering RoBERTa & \centering Llama & \centering FLAN & 97.29 & 0.968 & 0.022 & 0.946 \\
& \centering \textbf{RoBERTa} & \centering \textbf{-} & \centering \textbf{FLAN} & \textbf{97.77} & \textbf{0.978} & \textbf{0.022} & \textbf{0.955} \\
& \centering - & \centering GPT2 & \centering FLAN & 97.40 & 0.972 & 0.024 & 0.948 \\
& \centering - & \centering Llama & \centering FLAN & 96.47 & 0.993& 0.064& 0.931\\

\hline

\multirow{9}{*}{GPT} & \centering RoBERTa & \centering - & \centering - & 93.78 & 0.934 & 0.058 & 0.875 \\
& \centering - & \centering GPT2 & \centering - & 93.80 & 0.919 & 0.043 & 0.877 \\
& \centering - & \centering Llama & \centering - & 96.51 & 0.959 & 0.028 & 0.930 \\
& \centering - & \centering - & \centering FLAN &98.14 & 0.980 & 0.017 & 0.963 \\
& \centering RoBERTa & \centering GPT2 & \centering - & 97.23 & 0.975 & 0.031 & 0.945 \\
& \centering RoBERTa & \centering Llama & \centering - & 95.93 & 0.993 & 0.075 & 0.921 \\
& \centering RoBERTa & \centering GPT2 & \centering FLAN & 98.23 & 0.978 & 0.014 & 0.965 \\
& \centering RoBERTa & \centering Llama & \centering FLAN & 97.87 & 0.979 & 0.022 & 0.957 \\
& \centering \textbf{RoBERTa} & \centering \textbf{-} & \centering \textbf{FLAN} & \textbf{98.73} & \textbf{0.987}& \textbf{0.012} & \textbf{0.975}\\
& \centering - & \centering GPT2 & \centering FLAN & 98.25 & 0.970 & 0.005 & 0.965\\
& \centering - & \centering Llama & \centering FLAN & 97.17 & 0.956 & 0.012 & 0.944 \\
\hline

\multirow{9}{*}{Llama} & \centering RoBERTa & \centering - & \centering - &90.68 & 0.861 & 0.047 & 0.817 \\
& \centering - & \centering GPT2 & \centering - & 91.40 & 0.912 & 0.084 & 0.828 \\
& \centering - & \centering Llama & \centering - & 93.50 & 0.951 & 0.082 & 0.870 \\
& \centering - & \centering - & \centering FLAN & 94.92 & 0.961 & 0.063 & 0.899 \\
& \centering RoBERTa & \centering GPT2 & \centering - & 94.57 & 0.944 & 0.053 & 0.891 \\
& \centering RoBERTa & \centering Llama & \centering - & 95.61 & 0.952 & 0.040 & 0.912 \\
& \centering RoBERTa & \centering GPT2 & \centering FLAN & 93.10 & 0.972 & 0.111 & 0.865 \\
& \centering RoBERTa & \centering Llama & \centering FLAN & 95.38 & 0.970 & 0.062 & 0.908 \\
& \centering \textbf{RoBERTa} & \centering \textbf{-} & \centering \textbf{FLAN} & \textbf{96.53} & \textbf{0.949} & \textbf{0.018} & \textbf{0.931} \\
& \centering - & \centering GPT2 & \centering FLAN & 95.62 & 0.973 & 0.061 & 0.913 \\
& \centering - & \centering Llama & \centering FLAN & 95.22 & 0.941 & 0.037 & 0.905\\
\hline

\multirow{9}{*}{OPT} & \centering RoBERTa & \centering - & \centering - & 94.32 & 0.916 & 0.030 & 0.887 \\
& \centering - & \centering GPT2 & \centering - & 90.42 & 0.909 & 0.104 & 0.805 \\
& \centering - & \centering Llama & \centering - &95.86 & 0.970 & 0.053 & 0.917  \\
& \centering - & \centering - & \centering FLAN & 97.68 & 0.978 & 0.024 & 0.954 \\
& \centering RoBERTa & \centering GPT2 & \centering - & 95.59 & 0.982 & 0.071 & 0.913 \\
& \centering RoBERTa & \centering Llama & \centering - & 97.13 & 0.962 & 0.020 & 0.943 \\
& \centering RoBERTa & \centering GPT2 & \centering FLAN & 97.09 & 0.979 & 0.038 & 0.942  \\
& \centering RoBERTa & \centering Llama & \centering FLAN & 97.89 & 0.976 & 0.018 & 0.958 \\
& \centering \textbf{RoBERTa} & \centering \textbf{-} & \centering \textbf{FLAN} & \textbf{98.20} & \textbf{0.978} & \textbf{0.014} & \textbf{0.964} \\
& \centering - & \centering GPT2 & \centering FLAN & 97.70 & 0.987 & 0.033 & 0.954 \\
& \centering - & \centering Llama & \centering FLAN & 96.74 & 0.986 & 0.052 & 0.935\\
\hline

\multirow{5}{*}{Mistral} & \centering RoBERTa & \centering - & \centering - & 99.33 & 0.989 & 0.003 & 0.986 \\
& \centering - & \centering GPT2 & \centering - & 99.42 & 0.996 & 0.007 & 0.988 \\
& \centering - & \centering Llama & \centering - & 99.84 & 0.997 & 0.001 & 0.997 \\
& \centering - & \centering - & \centering FLAN & 99.72 & 0.997 & 0.003 & 0.994 \\
& \centering RoBERTa & \centering GPT2 & \centering - & 99.65& 0.999 & 0.005 & 0.993 \\
& \centering RoBERTa & \centering Llama & \centering - & 99.91 & 0.999 & 0.000 & 0.998 \\
& \centering RoBERTa & \centering GPT2 & \centering FLAN & 99.88 & 0.999 & 0.001 & 0.998 \\
& \centering RoBERTa & \centering Llama & \centering FLAN & 99.84 & 0.999 & 0.002 & 0.997 \\
& \centering \textbf{RoBERTa} & \centering \textbf{-} & \centering \textbf{FLAN} & \textbf{99.95} & \textbf{0.999} & \textbf{0.000} & \textbf{0.999} \\
& \centering - & \centering GPT2 & \centering FLAN & 99.86& 0.997 & 0.000 & 0.997 \\
& \centering - & \centering Llama & \centering FLAN & 98.84 & 1.000 & 0.023 & 0.977\\
\hline

\end{tabular}
\end{table*}
\subsection{Encoder-Decoder Language Models}

\begin{figure}[!ht]
    \centering
    \includegraphics[width=1\linewidth]{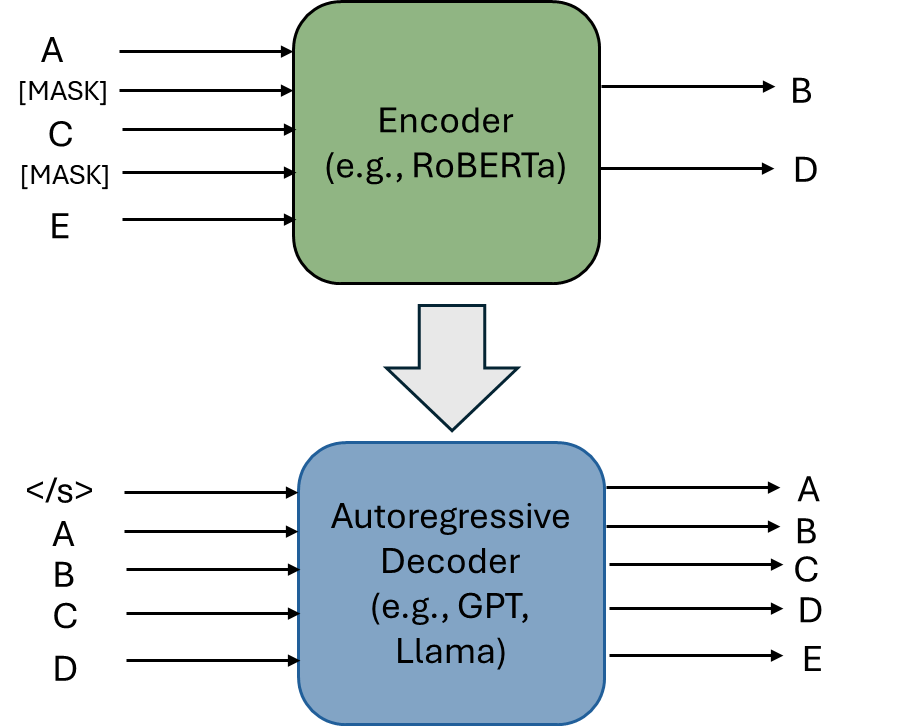}
    \caption{Encoder-Decoder Language Models transform an input sequence into a fixed-sized embedding. These embeddings are used by the decoder to generate an output sequence. Such models are often used in machine translation and sequence-to-sequence prediction.}
    \label{fig:encoder-decoder}
\end{figure}

The Encoder-Decoder model, as in Fig. \ref{fig:encoder-decoder}, serves as a versatile "text in, text out" architecture, proficient in generating a sequence of tokens \( y_1, ..., y_n \) based on an input sequence \( x_1, ..., x_m \). When presented with a sequence pair, the training objective revolves around maximizing the log-likelihood of  \( \log(P(y_1, ..., y_n | x_1, ..., x_m); \theta_T) \), where \( \theta_T \) represents the parameters within a complete encoder-decoder transformer model. To enrich the dataset for self-supervised pre-training, researchers explore various methods of sequence manipulation. These techniques involve altering the input token sequence in specific manners, with the objective of reconstructing the original sequence as the output. Examples of sequence corruption techniques include document rotation, sentence permutation, text infilling, and token deletion/masking, among others \cite{plm_survey}. FLAN \cite{flan-t5} demonstrates proficiency in capturing semantic correlations between input and output sequences, empowering it to generate translations or summaries that maintain coherence and contextual relevance.
\section{Observation and Results}
From Table \ref{results_63LLMs} and Table \ref{deepfake}, we observe that the embedding combination of MLMs and EDLMs performs the best on the authorship attribution task. In addition to classification accuracy, we have also documented the performance of our approach through the Matthews Correlation Coefficient (MCC). 

The equation for Matthews Correlation Coefficient (MCC) is provided in Eq. \ref{MCC_eq}. 
MCC assesses the correlation between the actual and predicted binary classifications, with values ranging from -1 to +1. A score of +1 indicates perfect prediction, 0 denotes no correlation, and -1 implies total disagreement between predictions and actual labels.

\begin{equation}
\label{MCC_eq}
MCC = \frac{TP \times TN - FP \times FN}{\sqrt{(TP + FP)(TP + FN)(TN + FP)(TN + FN)}}
\end{equation}
\vspace{0.05cm}

\begin{math}
    \hspace{1cm} TP - \text{True Positives} \hspace{2cm} TN - \text{True Negatives}
\end{math}

\begin{math}
    \hspace{1cm}FP - \text{False Positives}  \hspace{2cm} FN - \text{False Negatives}
\end{math}

\vspace{0.2cm}

Although classification accuracy is a common metric for assessing performance in binary classification tasks, it can sometimes be misleading. This is because the train and test split can introduce small biases, even in balanced datasets. Therefore, the MCC provides a more comprehensive view of performance as it accounts for all metrics: true positives, true negatives, false positives, and false negatives. Consequently, even a 0.01 increase in the MCC scores indicates a substantial improvement in performance.

As demonstrated in Table \ref{results_63LLMs} and Table \ref{results}, we notice that combining embeddings does not always enhance performance metrics. Incorporating embeddings from ALMs leads to a decrease in classification accuracy and MCC scores. This can be attributed to the nature of ALMs, which prioritise text generation and consequently assign more significance to the final tokens in the input text. 

\begin{table*}
\centering
\caption{Authorship attribution accuracies with different techniques for Deepfake text detection Dataset from the SQuAD domain.}
\label{results}
\begin{tabular}{|c|p{1.8cm}|p{1.5cm}|p{1.5cm}|c|c|c|c|}
\hline
\multirow{2}{*}{\textbf{Source}} &  \multicolumn{3}{|c|}{\textbf{Embedding Model}} & \multirow{2}{*}{\textbf{Accuracy (\%)}} & \multirow{2}{*}{\textbf{TPR}} & \multirow{2}{*}{\textbf{FPR}} & \multirow{2}{*}{\textbf{MCC}}\\
\cline{2-4}
 & \centering\textbf{Encoder} & \centering\textbf{Decoder} & \textbf{Encoder-Decoder} & & & & \\
 \hline

\multirow{10}{*}{FLAN} & \centering RoBERTa & \centering- & \centering- & 94.69 & 0.934 & 0.004 & 0.894 \\
& \centering - & \centering GPT2 & \centering - & 94.33 & 0.929 &  0.042 &  0.887 \\
& \centering - & \centering - & \centering FLAN & 97.54 & 0.967 & 0.020 & 0.948 \\

& \centering - & \centering Llama & \centering - & 98.56 & 0.978 & 0.007 & 0.971 \\
& \centering RoBERTa & \centering GPT2 & \centering - & 97.70 & 0.984 & 0.030 & 0.954 \\
& \centering RoBERTa & \centering Llama & \centering - & 99.01 & 0.993 & 0.013 & 0.980 \\
& \centering \textbf{RoBERTa} & \centering \textbf{-} & \centering \textbf{FLAN} & \textbf{99.45} & \textbf{0.995} & \textbf{0.006} & \textbf{0.989} \\
& \centering RoBERTa & \centering GPT2 & \centering FLAN & 98.87 & 0.993 & 0.017 & 0.976  \\
& \centering RoBERTa & \centering Llama & \centering FLAN & 99.23 & 0.991 & 0.006 & 0.985 \\
& \centering - & \centering Llama & \centering FLAN & 99.23 & 0.998 & 0.013 & 0.985 \\
& \centering - & \centering GPT2 & \centering FLAN & 99.38 & 0.998 & 0.010 & 0.988 \\
\hline

\multirow{10}{*}{GPT} & \centering RoBERTa & \centering - & \centering - & 94.69 & 0.934 & 0.04 & 0.894 \\
& \centering - & \centering GPT2 & \centering - & 88.88 & 0.868 & 0.093 & 0.776 \\
& \centering - & \centering - & \centering FLAN & 95.27 & 0.943 & 0.040 & 0.903 \\
& \centering - & \centering Llama & \centering - & 88.88 & 0.868 & 0.093 & 0.776 \\
& \centering RoBERTa & \centering GPT2 & \centering - & 95.29 & 0.948 & 0.042 & 0.906 \\
& \centering RoBERTa & \centering Llama & \centering - & 97.04 & 0.975 & 0.034 & 0.941 \\
& \centering \textbf{RoBERTa} & \centering \textbf{-} & \centering \textbf{FLAN} & \textbf{97.48} & \textbf{0.973} & \textbf{0.023} & \textbf{0.949} \\ 
& \centering RoBERTa & \centering GPT2 & \centering FLAN & 96.34 & 0.964 & 0.038 & 0.925 \\
& \centering RoBERTa & \centering Llama & \centering FLAN & 97.04 & 0.959 & 0.019 & 0.941\\
& \centering - & \centering Llama & \centering FLAN & 95.83 & 0.966 & 0.049 & 0.917 \\
& \centering - & \centering GPT2 & \centering FLAN & 96.71 & 0.948 & 0.015 & 0.935 \\
\hline

\multirow{10}{*}{Llama} & \centering RoBERTa & \centering - & \centering - &94.69 & 0.934 & 0.04 & 0.894 \\
& \centering - & \centering GPT2 & \centering - & 93.99 & 0.943 & 0.065 & 0.878 \\
& \centering - & \centering - & \centering FLAN & 95.86 & 0.992  & 0.076 & 0.918 \\
& \centering - & \centering Llama & \centering - & 94.52 & 0.940 & 0.050 & 0.890 \\
& \centering RoBERTa & \centering GPT2 & \centering - & 97.61 & 0.975 & 0.023 & 0.952 \\
& \centering RoBERTa & \centering Llama & \centering - & 98.25 & 0.977 & 0.013 & 0.965 \\
& \centering \textbf{RoBERTa} & \centering \textbf{-} & \centering \textbf{FLAN} &  \textbf{98.90} & \textbf{0.982} & \textbf{0.004} & \textbf{0.978} \\
& \centering RoBERTa & \centering GPT2 & \centering FLAN & 98.79 & 0.982 & 0.008 & 0.974  \\
& \centering RoBERTa & \centering Llama & \centering FLAN & 98.79 & 0.977 & 0.002 & 0.976\\
& \centering - & \centering Llama & \centering FLAN & 98.45 & 0.013  & 0.982 & 0.969 \\
& \centering - & \centering GPT2 & \centering FLAN & 98.58 & 0.977 & 0.005 & 0.972 \\
\hline

\multirow{5}{*}{OPT} & \centering RoBERTa & \centering - & \centering - & 94.69 & 0.934 & 0.04 & 0.894 \\
& \centering - & \centering GPT2 & \centering - & 91.77 & 0.900 & 0.067 & 0.834 \\
& \centering - & \centering - & \centering FLAN & 97.77 & 0.967 & 0.017 & 0.951  \\
& \centering - & \centering Llama & \centering - &97.41 & 0.976 & 0.028 & 0.948 \\
& \centering RoBERTa & \centering GPT2 & \centering - & 95.72 & 0.961 & 0.047 & 0.914\\
& \centering RoBERTa & \centering Llama & \centering - & 96.38 & 0.966 & 0.038 & 0.928\\
& \centering \textbf{RoBERTa} & \centering \textbf{-} & \centering \textbf{FLAN} & \textbf{98.57} & \textbf{0.975} & \textbf{0.004} & \textbf{0.972} \\
& \centering RoBERTa & \centering GPT2 & \centering FLAN & 97.33  & 0.964 & 0.019 & 0.945 \\
& \centering RoBERTa & \centering Llama & \centering FLAN & 97.15 & 0.954 & 0.013 & 0.943\\
& \centering - & \centering Llama & \centering FLAN & 97.79 & 0.975 & 0.020 & 0.956 \\
& \centering - & \centering GPT2 & \centering FLAN & 98.10 & 0.977 & 0.015 & 0.962 \\
\hline

\end{tabular}
\label{deepfake}
\end{table*}

\section{Conclusion and Future Work}

This paper presents a pioneering framework that employs Embedding Fusion to address the longstanding challenge of distinguishing between AI-generated and human-authored text. Our approach integrates embeddings from Masked Language Models (MLMs) and Encoder-Decoder Language Models (EDLMs), concatenating them into a single feature vector. This vector is subsequently reshaped into a 2D representation to enhance the capture of inter-embedding relationships. Through extensive experimentation across 2 datasets, we achieve an accuracy >96\% and Matthews Correlation Coefficient (MCC) score  >0.93 showcases its effectiveness. Moreover, our findings indicate that incorporating embeddings from Autoregressive Language Models (ALMs) can degrade the information within the feature vector. We believe that the embedding fusion methodology holds significant potential for advancing authorship attribution tasks, with opportunities for further exploration through attention mechanisms and interleaving strategies.

\bibliographystyle{splncs04}
\bibliography{egbib, llm_references}

\end{document}